\documentclass[letterpaper]{article} 
\usepackage{aaai25}  
\usepackage{times}  
\usepackage{helvet}  
\usepackage{courier}  
\usepackage[hyphens]{url}  
\usepackage{graphicx} 
\usepackage{natbib}  
\usepackage{caption} 
\usepackage{algorithm}
\usepackage{algorithmic}
\usepackage{amsmath}
\usepackage{amsfonts}
\usepackage{amssymb}
\usepackage{pifont}
\usepackage{booktabs}
\usepackage{etoolbox}
\usepackage{xcolor}
\nocopyright
\usepackage{newfloat}
\usepackage{listings}
\DeclareCaptionStyle{ruled}{labelfont=normalfont,labelsep=colon,strut=off} 
\floatstyle{ruled}
\newfloat{listing}{tb}{lst}{}
\floatname{listing}{Listing}
\title{DiffuseHigh: Training-free Progressive High-Resolution Image Synthesis through Structure Guidance}
\author {
    Younghyun Kim\equalcontrib\textsuperscript{\rm 1},
    Geunmin Hwang\equalcontrib\textsuperscript{\rm 1},
    Junyu Zhang\textsuperscript{\rm 2,3},
    Eunbyung Park\textsuperscript{\rm 1,3}\thanks{Corresponding author.}
}
\affiliations {
    \textsuperscript{\rm 1}Department of Artificial Intelligence, Sungkyunkwan University\\
    \textsuperscript{\rm 2}School of Computer Science and Engineering, Central South University\\
    \textsuperscript{\rm 3}Department of Electrical and Computer Engineering, Sungkyunkwan University\\
}

\begin{document}
\maketitle


\begin{figure*}[ht!]
    \centering
    \includegraphics[width=1.0\textwidth]{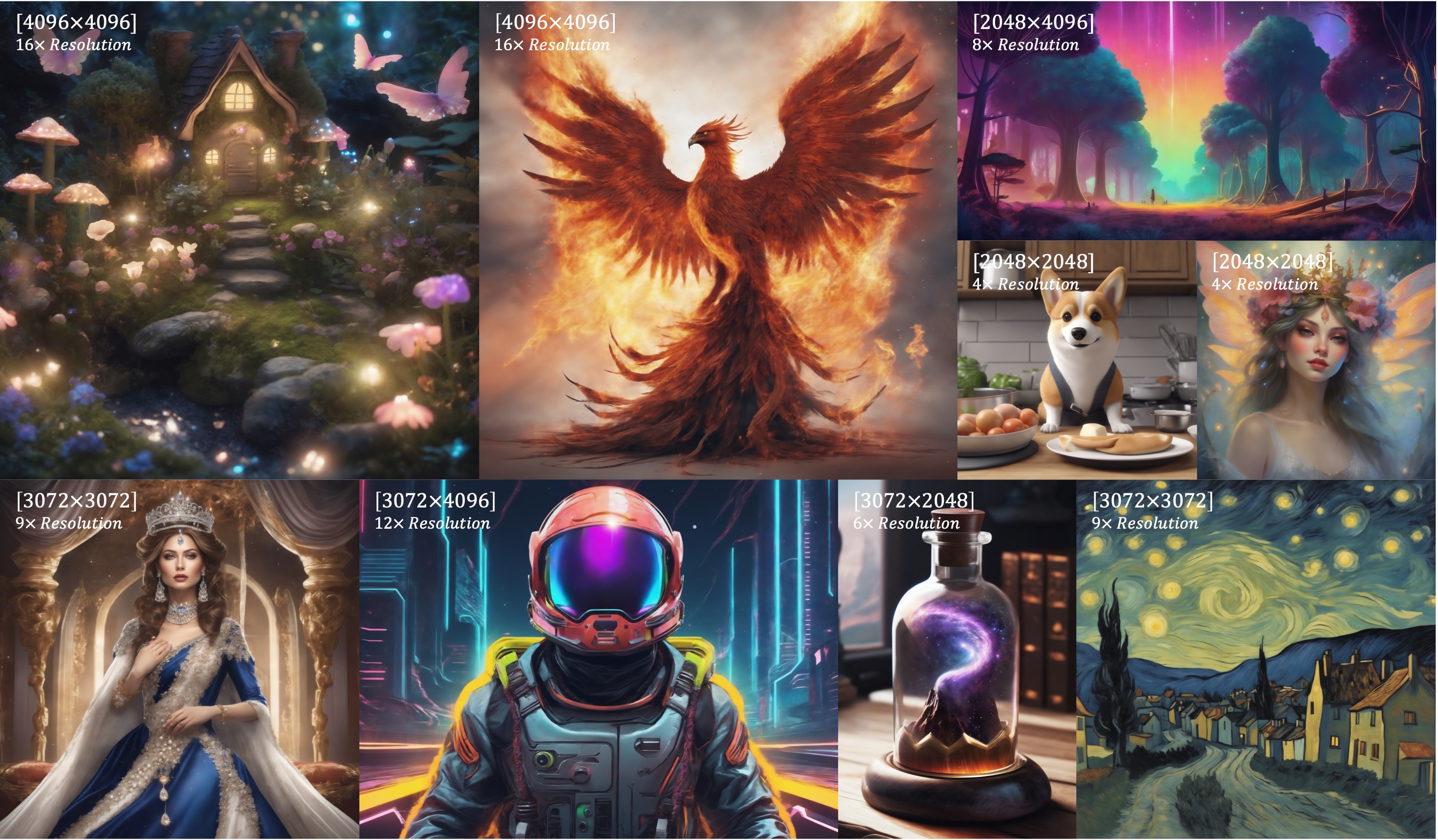} 
    \caption{\textbf{Qualitative examples of the proposed DiffuseHigh pipeline.} \textit{DiffuseHigh} enables the pre-trained text-to-image diffusion models (SDXL in this figure) to generate higher-resolution images than the originally trained resolution, e.g., 4$\times$, 16$\times$, without any training or fine-tuning.}
    \label{fig:figure_main}
    \vspace{-10pt}
\end{figure*}

\begin{abstract}
Large-scale generative models, such as text-to-image diffusion models, have garnered widespread attention across diverse domains due to their creative and high-fidelity image generation. Nonetheless, existing large-scale diffusion models are confined to generating images of up to 1K resolution, which is far from meeting the demands of contemporary commercial applications. Directly sampling higher-resolution images often yields results marred by artifacts such as object repetition and distorted shapes. Addressing the aforementioned issues typically necessitates training or fine-tuning models on higher-resolution datasets. However, this poses a formidable challenge due to the difficulty in collecting large-scale high-resolution images and substantial computational resources. While several preceding works have proposed alternatives to bypass the cumbersome training process, they often fail to produce convincing results. In this work, we probe the generative ability of diffusion models at higher resolution beyond their original capability and propose a novel progressive approach that fully utilizes generated low-resolution images to guide the generation of higher-resolution images. Our method obviates the need for additional training or fine-tuning which significantly lowers the burden of computational costs. Extensive experiments and results validate the efficiency and efficacy of our method.
\begin{links}
    \link{Project Page}{https://yhyun225.github.io/DiffuseHigh}
\end{links}

\end{abstract}

\section{Introduction}
With the establishment of diffusion models, there have been rapid advancements across various domains, including audio synthesis~\cite{kong2020diffwave,chen2020wavegrad,lam2021bilateral,liu2023audioldm}, image synthesis~\cite{ho2020denoising,song2020denoising,dhariwal2021diffusion,gao2023masked}, video generation~\cite{he2022latent,ho2022imagen,blattmann2023align,wang2023modelscope,blattmann2023stable,chen2023videocrafter1}, and 3D generation~\cite{poole2022dreamfusion,wang2024prolificdreamer,lin2023magic3d,chen2023fantasia3d,shi2023mvdream,tang2023dreamgaussian,yi2023gaussiandreamer}. Notably, text-to-image diffusion models~\cite{balaji2022ediffi,rombach2022high,podell2023sdxl,saharia2022photorealistic,ramesh2022hierarchical} have attracted considerable attention due to their ability to generate visually captivating images using intuitive, human-friendly natural language descriptions. Stable Diffusion (SD) and Stable Diffusion XL (SDXL), the open-source text-to-image diffusion models trained on a large-scale online dataset~\cite{schuhmann2022laion}, have emerged as prominent tools for a diverse range of generative tasks. These tasks include but are not limited to image editing~\cite{avrahami2022blended,hertz2022prompt,tumanyan2023plug,kawar2023imagic}, inpainting~\cite{rombach2022high,saharia2022palette,lugmayr2022repaint}, super-resolution~\cite{rombach2022high,saharia2022image,gao2023implicit}, and image-to-image translation~\cite{brooks2023instructpix2pix,mou2023t2i,yu2023freedom,zhang2023adding}.

Despite the promising performance exhibited by SD and SDXL, they encounter limitations when generating images at higher-resolutions beyond their training resolution. The direct inference of unseen high-resolution samples often reveals repetitive patterns and irregular structures, particularly noticeable in object-centric samples, as discussed in prior works~\cite{he2023scalecrafter,du2024demofusion}. While a straightforward approach might involve training or fine-tuning diffusion models on higher-resolution images, several challenges impede this approach. First, collecting text-image pairs of higher-resolution is not readily feasible. Second, training on large-resolution images demands substantial computational resources due to the increased size of the intermediate features. Furthermore, capturing and learning the features from high-dimensional data often requires a greater model capacity (more model parameters), leading to further computational strain on the training process.

Several tuning-free~\cite{bar2023multidiffusion,lee2024syncdiffusion,he2023scalecrafter,du2024demofusion} methods proposed various approaches to adapt pre-trained diffusion models to higher resolutions beyond their original settings. MultiDiffusion~\cite{bar2023multidiffusion} and SyncDiffusion~\cite{lee2024syncdiffusion} employ joint diffusion processes with overlapping windows, each corresponding to different region within the generating image. These models can produce images of arbitrary shape, but the resulting image involves object repetition issues since the non-overlapping patches do not correlate to each other, lacking perception of global context during the denoising process. ScaleCrafter~\cite{he2023scalecrafter}, on the other hand, extends the receptive field of the diffusion model by dilating the pre-trained convolution weights of the denoising UNet~\cite{ronneberger2015u}. While it effectively addresses repetition issues in certain instances, its success heavily depends on the extensive search of the hyperparameters. 

In this work, we investigate the text-to-image diffusion model's capability of generating previously unseen high-resolution images and introduce a novel approach that does not involve any training (or fine-tuning) and additional modules. Moreover, our proposed method does not modify the pre-trained weights or the architecture of the denoising network, which eliminates the labor of searching for the optimal hyperparameters involved in the pipeline, and is more robust to certain hyperparameters. We posit that text-to-image diffusion models trained on internet-scale datasets innately possess the potential to generate images at resolutions higher than their training resolution thanks to its convolutional architecture~\cite{rombach2022high} and broad data distribution coverage.

We introduce a novel progressive high-resolution image generation pipeline, dubbed \textit{DiffuseHigh}, where relatively low-resolution (training-resolution of the pre-trained diffusion models) images serve as structural guidance for generating higher-resolution images. Inspired by the recent literature~\cite{meng2021sdedit, podell2023sdxl, guo2024make}, our proposed pipeline involves a noising-denoising loop to synthesize higher-resolution images. First, we generate the low-resolution image using the base diffusion model and upsample it with arbitrary interpolation, e.g., bilinear interpolation. Then, we add sufficient noise to obfuscate the fine details of the interpolated images. Finally, we perform the reverse diffusion process to denoise those images to infuse the high-frequency details to synthesize higher-resolution images and repeat this process until we obtain the desired resolution images. This approach leverages the overall structure from the low-resolution image, effectively addressing repetition issues observed in the prior methods.

However, the `adding noise to damage the images' approach poses several challenges. If we add too much noise, then we lose most of the structure in the low-resolution images, resulting in repetitive outcomes similar to those we generate from scratch. On the other hand, if we introduce a minimal amount of noise, the generated higher-resolution images do not show notable differences from the interpolated images, losing the opportunity to synthesize high-frequency details. In addition, finding adequate noise relies on both the content of the image and the pre-trained models, which makes it challenging to offer precise suggestions to users. 

To resolve the issues above, we propose a principled way of preserving the overall structure from the low-resolution image for the suggested progressive pipeline. We employ a frequency-domain representation to extract the global structure as well as detailed contents from the low-resolution images. More specifically, we adopt the Discrete Wavelet Transform (DWT) to obtain essential contents, e.g., the approximation coefficient, which we then incorporate into the denoising procedure to ensure that the resulting image remains consistent and does not deviate excessively. Fig.~\ref{fig:fig_pipeline} provides an overview of the overall pipeline of our method. 

The contributions of our work are summarized as follows:
\begin{itemize}
    
    \item We suggest a novel training-free progressive high-resolution image synthesis pipeline called \textit{DiffuseHigh}, in which a lower-resolution image acts as a guide for generating higher-resolution images. 
    
    \item Our proposed method involves Discrete Wavelet Transform (DWT)-based structure guidance during the denoising process, which enhances both the structural properties and fine details of the generated samples.

    \item We conduct comprehensive experiments and ablation studies on high-resolution image synthesis, demonstrating the superiority and versatility of our method.
    
\end{itemize}


\section{Related Works}
\subsubsection{Text-to-Image Generation}
Recently, diffusion models (DMs) have gained popularity for their ability to produce high-quality images~\cite{peebles2023scalable}, showcasing great potential in text-to-image generation~\cite{nichol2021glide,ho2022imagen,ramesh2022hierarchical}.
Especially the pioneering work, Stable Diffusion~\cite{rombach2022high} and Stable Diffusion XL~\cite{podell2023sdxl} have garnered broad attention due to their astonishing image quality and computational efficiency.
Moreover, thanks to their large-scale training, they have been applied to various text-to-image tasks~\cite{li2024snapfusion, nichol2021glide, chang2023muse} by fine-tuning~\cite{ruiz2023dreambooth} or using training-free~\cite{ramesh2021zero} methods.

\subsubsection{High-resolution Image Synthesis}
Despite advancements in diffusion-based image synthesis methods, achieving high-resolution image generation remains elusive. Direct inference of SD and SDXL produces samples with repetitive patterns and irregular structures~\cite{he2023scalecrafter}. Previous studies have tackled these challenges through training from scratch or fine-tuning~\cite{xie2023difffit, zheng2023any, guo2024make}. However, these methods often necessitate substantial computational resources and considerable amount of high-resolution text-paired training dataset. Consequently, there is a growing trend towards training-free methods for generating high-resolution images.

ScaleCrafter~\cite{he2023scalecrafter} employs dilated convolution to modify the receptive field of convolutions in denoising UNet, enabling high-resolution image generation without the need for training. FouriScale~\cite{huang2024fouriscale} further incorporate a low-pass operation, which improves structural and scale consistency. HiDiffusion~\cite{zhang2023hidiffusion} identifies that the object repetition problem primarily originates from the deep blocks in the denoising UNet and proposes alternative UNet which dynamically adjust the feature map size during the denoising process. Additionally, they successively reduce the computational burden by modifying the self-attention blocks of the UNet. However, we argue that modifying the weights or the architecture of the pre-trained diffusion model has risk of degrading the model performance, often resulting in undesirable deformations in images (See Fig.\ref{fig:fig_comparison_with_baselines}). DemoFusion~\cite{du2024demofusion} leverages skip residual connections and dilated sampling to generate higher-resolution images in a progressive manner. Despite their efforts, it suffers from the irregular patterns and repetition of small objects in localized areas of the result images, and also from the slow generation speed. AccDiffusion~\cite{lin2024accdiffusion} addresses these issues with patch-wise prompt and improved dilated sampling, but still suffers from extremely slow inference speed.

Concurrently, ResMaster~\cite{shi2024resmaster} also proposed an algorithm that leverages the low-frequency information of the latent of the guidance image, in order to provide desirable global semantics during the denoising process. Different from theirs, we explicitly obtain structural guidance from the reconstructed image using DWT.

\section{Method}
\label{sec:method}
Our work aims to generate higher-resolution images over training size given textual prompts with a text-to-image diffusion models in a training-free manner. In this work, we mainly utilize SDXL~\cite{podell2023sdxl} as our base model. We provide preliminaries related to our work in the appendix.

\subsection{Problem Formulation}
Given a text description $y$ and SDXL $D_{\phi}(\cdot)$ trained on fixed-size images $x^L_0 \in \mathbb{R}^{h \times w \times 3}$, our objective is to generate higher resolution images $x^H_0 \in \mathbb{R}^{H \times W \times 3}$ without training or modifying $\phi$, where $h \ll H, w \ll W$.

\subsection{Progressive High-Resolution Image Generation}
We first present the progressive high-resolution image generation strategy of \textit{DiffuseHigh}, equipped with a pretrained SDXL model. Initially, given text prompt $y$, our method starts with a clean image $x_0^L \in \mathbb{R}^{h \times w \times 3}$, either generated by the SDXL or provided by the user. Assuming alignment between the generated image and the provided text, we employ arbitrary interpolation, e.g., bilinear interpolation, to upscale the image:
\begin{equation}
    \tilde{x}_0^H = \texttt{INTERP}(x_0^L) \in \mathbb{R}^{H \times W \times 3}
    \label{eq:eq1}
\end{equation}
Note that the details of $\tilde{x}_0^H$ lack clarity due to the nature of the interpolation, which entails averaging neighboring pixel values to compose newly introduced pixels.

\begin{figure*}[ht!]
    \centering
    \includegraphics[width=1.0\textwidth]{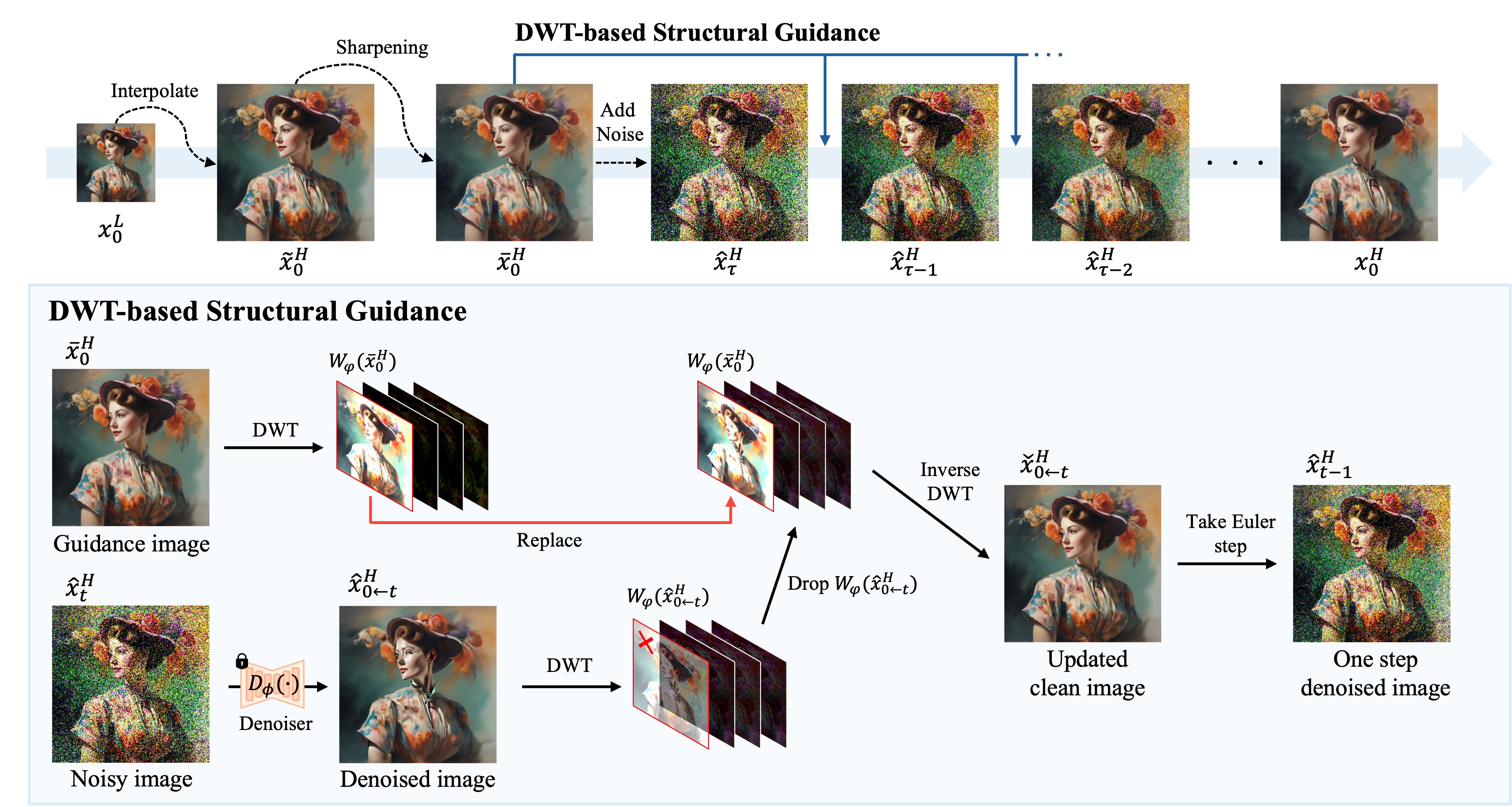} 
    \caption{\textbf{Progressive High-Resolution \textit{DiffuseHigh} Pipeline.} Overall pipeline of our proposed \textit{DiffuseHigh}. For simplicity, we did not depict transformation between latent space and pixel space.}
    \label{fig:fig_pipeline}
    \vspace{-10pt}
\end{figure*}

In order to infuse the appropriate details into $\tilde{x}_0^H$, we first add noise corresponding to the diffusion timestep $\tau < T$ to its latent code $\tilde{z}_0^H = \mathcal{E}(\tilde{x}_0^H)$:
\begin{equation}
    \hat{z}_\tau^H = \tilde{z}_0^H + \epsilon, \quad \epsilon \sim \mathcal{N}(0, \sigma_\tau^2 I),
    \label{eq:eq2}
\end{equation}
where $\sigma_\tau^2$ is the variance of the Gaussian noise at timestep $\tau$. We selected the noising diffusion timestep $\tau$ where the noisy image reconstructed from the latent decoder $\mathcal{D}(\cdot)$, $\hat{x}_\tau^H = \mathcal{D}(\hat{z}_\tau^H)$, preserves the global structures. Then the denoising network $D_{\phi}(\cdot)$ performs the iterative reverse process on the noisy latent representation $\hat{z}_\tau^H$ to recover the clean latent $z_0^H$. Finally, we obtain the desired high-resolution image $x_0^H \in \mathbb{R}^{H \times W \times 3}$ by employing the latent decoder, i.e., $x_0^H = \mathcal{D}(z_0^H)$. We repeat this process iteratively until we obtain the desired higher-resolution image.

The noising-denoising technique adopted in our work gradually projects the sample onto the manifold of natural, highly detailed images that the diffusion model has learned. As shown in Make-a-Cheap-Scaling~\cite{guo2024make}, this process enables the injection of high-frequency details into the interpolated high-resolution image. Nonetheless, we observed numerous instances where solely applying this simple approach degraded the image quality, typically suffering from repeated small objects or deformed local details in the image. This lead us to develop a more principled way to uphold the overall structure and maintain the quality of the generated higher-resolution images.

\subsection{Structural Guidance through DWT}
\label{subsec:structural_guidance}
To remedy the aforementioned issues, we hereby introduce a \textbf{structural guidance} by incorporating a Discrete Wavelet Transform (DWT). This method aims to enhance the fidelity of generated images by encouraging the preservation of crucial features from the low-resolution input.

Let $\varphi$ be the two-dimensional scaling function, and $\psi^H$, $\psi^V$, $\psi^D$ the two-dimensional wavelets, each corresponding to the horizontal (H), vertical (V), and diagonal directions (D), respectively. Then, the single level 2D-DWT decomposition of the image $x$ can be written as follows:
\begin{equation}
    \texttt{DWT}(x) := \{W_\varphi(x)\} \cup \{W_{\psi^i}(x)\}_{i \in \{H,V,D\}},
\end{equation}
where $W_\varphi(x)$ the approximation coefficient, and $W_{\psi^i}(x)$ the detail coefficients along the direction $i \in \{H, V, D\}$.

 Considering that $W_\varphi(x)$ contains the global features of the image $x$, given an interpolated image $\tilde{x}_0^H \in \mathbb{R}^{H \times W \times 3}$ obtained from Eq. (\ref{eq:eq1}), we extract its low-frequency component $W_\varphi(\tilde{x}_0^H)$ utilizing the DWT, which encapsulates the overall structure and coarse details of the image. Then, during the progressive denoising process, we replace the low-frequency component of the estimated clean image $\hat{x}_{0 \leftarrow t}^H = \mathcal{D}(\hat{z}_{0 \leftarrow t}^H)$, with the extracted low-frequency component at timestep `$t$' as follows:
\begin{equation}
    \check{x}_{0 \leftarrow t}^H = \texttt{iDWT}(\{W_\varphi(\tilde{x}_0^H)\} \cup \{W_{\psi^i}(\hat{x}_{0 \leftarrow t}^H)\}_{i \in \{H,V,D\}})
    \label{eq:eq3}
\end{equation}

where $\hat{z}_{0 \leftarrow t}^H = D_\phi(\hat{z}_t^H; \sigma_t)$ and $\texttt{iDWT}$ denotes inverse DWT. Then, the updated estimated clean image $\check{x}_{0 \leftarrow t}^H$ is encoded back into the latent space to sample the next noisy latent.

\begin{figure}[ht!]
    \centering
    \includegraphics[width=0.47\textwidth]{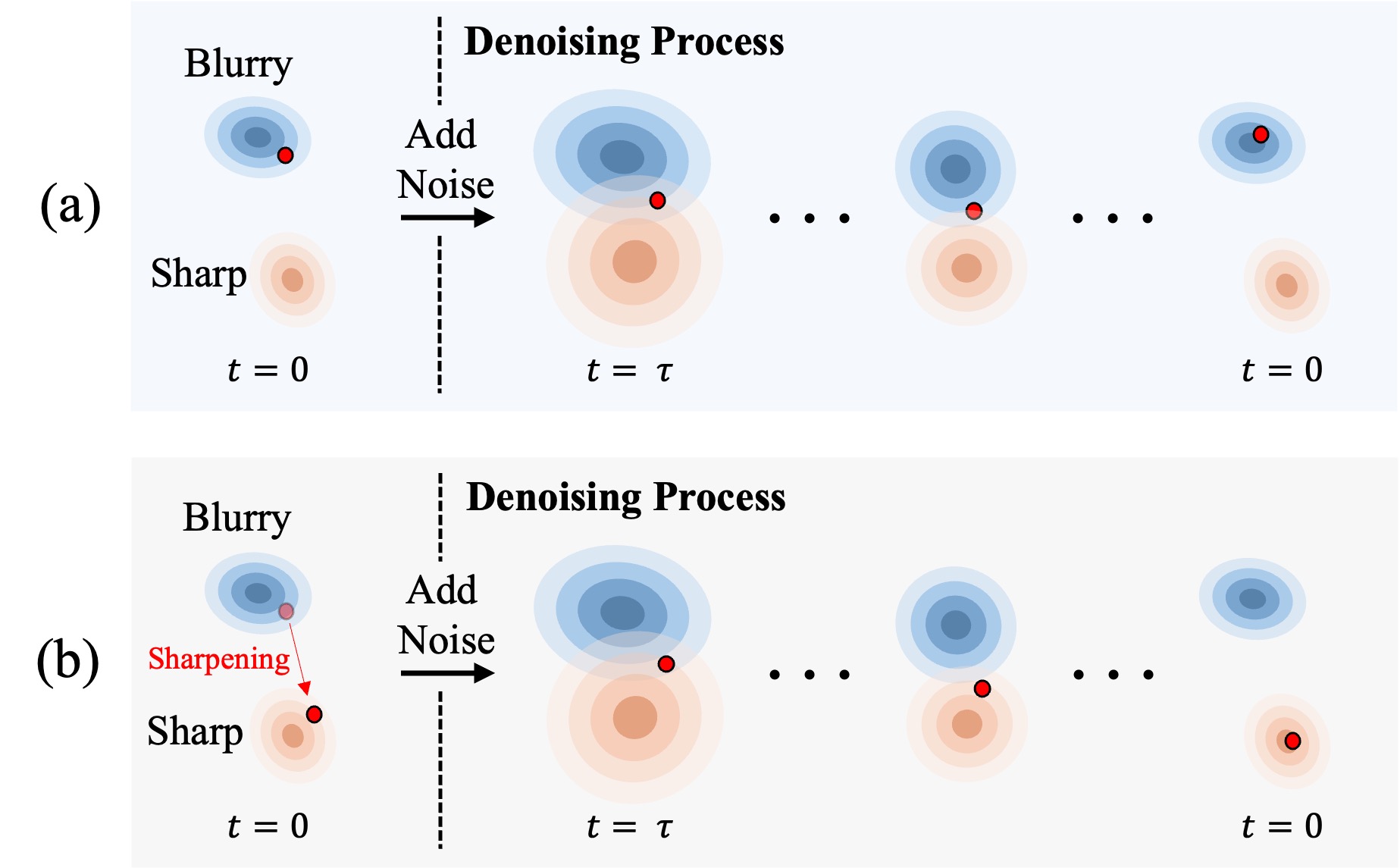} 
    \caption{\textbf{Data sample toward sharp data distribution mode with sharpening.} (a) Without sharpening, (b) With sharpening. Red dot represents the data point. We encourage data point to move toward the sharp data distribution mode during denoising process by sharpening the blurry image.}
    \label{fig:fig_sharpening_explanation}
    \vspace{-20pt}
\end{figure}

Previous studies~\cite{ho2020denoising,rombach2022high} present that the reverse process over each timestep performs denoising on different \textbf{levels} of the image, from semantic to perceptual, or low-frequency to high frequency details. Since the global structures and low frequency details are fixed and barely changed at the latter part of the denoising process, we found it beneficial to apply our structural guidance only at the early stages of our denoising process. Furthermore, this strategy significantly lowers the computational burden of our pipeline since our pipeline acquires low-frequency guidance from the reconstructed image, which requires frequent transition between the latent space and the pixel space. Empirically, we found that applying structural guidance steps $\delta = 5$ out of $\tau = 15$ steps yields the best results.

\subsection{Boosting the Image Quality with Sharpening}
Our proposed structural guidance effectively transfers the correct global context from a low-resolution image to a high-resolution image, maintaining the global coherence of the image. However, the generated image often appears blurry with smooth textures. We hypothesize that this blurriness arises from the interpolation used in our pipeline for the following reasons: (1) It is apparent that the diffusion models trained on large-scale datasets have the prior of blurry samples. Adding noise to the interpolated image, which lies near the blurry data distribution mode, is more likely to result in a blurry image after the denoising process (Fig.~\ref{fig:fig_sharpening_explanation} (a)). (2) Interpolating a low-resolution image involves averaging neighboring pixel values, thus creating smooth transitions between pixels. These low intensity changes in object boundaries and edges are easily incorporated into the low-frequency information and subsequently transferred to the target image through our DWT-based structural guidance.

To address the blurriness issue, we apply the sharpening operation to the interpolated image $\tilde{x}_0^H$:
\begin{equation}
    \bar{x}_{0}^H = (\alpha + 1)\tilde{x}_0^H - \alpha\mathcal{S}(\tilde{x}_0^H)
\end{equation}  
where $\mathcal{S}$ is an arbitrary smoothing operation and $\alpha$ is the sharpness factor that controls the magnitude of the sharpness. This behavior slightly moves the sample point closer to the sharp data distribution mode, resulting in a sharp and clear sample after the denoising process (Fig.~\ref{fig:fig_sharpening_explanation} (b)), and also causes meaningful intensity changes at edges and boundaries of the interpolated image. Surprisingly, we found that simply sharpening the image significantly alleviates the aforementioned issues. We further provide extensive analysis on this phenomenon in the appendix. The overall pipeline of \textit{DiffuseHigh} is illustrated in Fig.~\ref{fig:fig_pipeline}.

\begin{figure*}[ht!]
    \centering
    \includegraphics[width=1.0\textwidth]{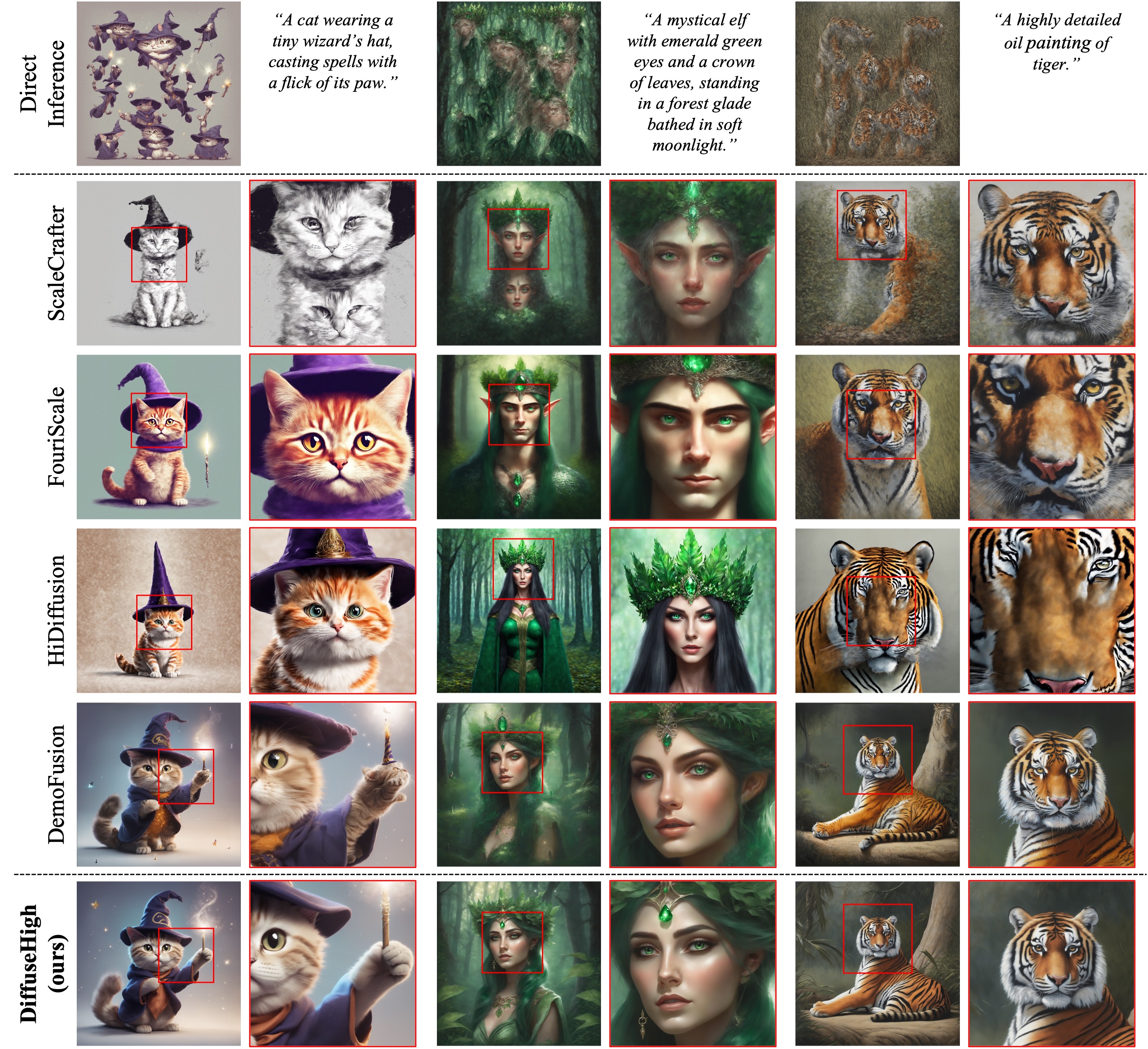} 
    \caption{\textbf{Qualitative comparison to baselines in 4096 $\times$ 4096 resolution experiment.} Please \textbf{ZOOM-IN} the figure in order to see the details of each image.} 
    \label{fig:fig_comparison_with_baselines}
    \vspace{-10pt}
\end{figure*}

\begin{table*}[ht!]
    \centering
    \resizebox{\textwidth}{!}{
        \begin{tabular}{c|c c c c c |c|c c c c c |c|c c c c c |c}
            \toprule
                 & \multicolumn{6}{c|}{$2048 \times 2048$} & \multicolumn{6}{c|}{$2048 \times 4096$} & \multicolumn{6}{c}{$4096 \times 4096$} \\ \hline
                
               Methods & $\text{FID}_r (\downarrow)$ & $\text{KID}_r (\downarrow)$ & $\text{FID}_p (\downarrow)$ & $\text{KID}_p (\downarrow)$ & CLIP $(\uparrow)$ & Time & $\text{FID}_r (\downarrow)$ & $\text{KID}_r (\downarrow)$ & $\text{FID}_p (\downarrow)$ & $\text{KID}_p (\downarrow)$ & CLIP $(\uparrow)$ & Time & $\text{FID}_r$ & $\text{KID}_r (\downarrow)$ & $\text{FID}_p (\downarrow)$ & $\text{KID}_p (\downarrow)$ & CLIP $(\uparrow)$ & Time \\

            \midrule
               Direct Inference & 94.11 & 0.021 & 66.72 & 0.020 & 29.96 & 42 sec & 139.44& 0.052 & 79.67 & 0.026 & 27.82 & 104 sec & 173.70 & 0.067 & 92.42 & 0.032 & 25.12 & 329 sec\\
               
               ScaleCrafter & 79.95 & 0.015 & 59.32 & 0.016 & 29.56 & 41 sec & 132.43& 0.044 & 114.94 & 0.055 & 25.42 & 189 sec & 112.81 & 0.031 & 108.63 & 0.047 & 27.07 & 585 sec\\
               
               FouriScale & 67.05 & 0.010 & 51.13 & 0.013& 30.85 & 73 sec& 117.52 & 0.036 & 114.93 & 0.055 & 26.49 & 194 sec & 97.40 & 0.023 & 104.62 & 0.041 & 28.37 & 563 sec\\
               
               HiDiffusion & 75.59 & 0.011 & 54.77 & 0.013 & 29.17 & 24 sec& 102.79& 0.026 & 81.68 & 0.027 & 26.68 & 58 sec & 127.55 & 0.044 & 153.98 & 0.081 & 24.44 & 105 sec\\
               
               DemoFusion & \underline{57.16} & \textbf{0.007} & \underline{35.54} & \underline{0.010} & \textbf{31.95} & 94 sec & \underline{63.56}& \underline{0.010} & \underline{49.08} & \underline{0.017}& \underline{29.13} & 319 sec & \underline{62.82} & \underline{0.010} & \underline{48.97} & \underline{0.017} & \textbf{31.59} & 728 sec\\
            \bottomrule
               DiffuseHigh & \textbf{56.90} & \underline{0.008} & \textbf{34.14} & \textbf{0.009} & \underline{31.26} & 43 sec & \textbf{53.83}& \textbf{0.007}& \textbf{33.73}& \textbf{0.009}&  \textbf{30.50} & 78 sec &\textbf{56.09} & \textbf{0.007} & \textbf{38.93} & \textbf{0.010} & \underline{31.32} & 258 sec \\
            \hline
        \end{tabular}}
    \caption{\textbf{Quantitative results of higher-resolution image generation experiments}. Hereinafter, we represent the best results with \textbf{bold} and second best with \underline{underline}. We measured the inference time of each method by averaging the time generating 10 images in a single NVIDIA A100 gpu.}
    \label{tab:tab_main}
    \vspace{-10pt}
\end{table*}


\section{Experiments}
In this section, we report the qualitative and quantitative results of our proposed \textit{DiffuseHigh}. We also provide extensive ablation studies to validate the efficacy of our method thoroughly.

\subsection{Implementation Details}
We mainly conducted our experiments with SDXL~\cite{podell2023sdxl}, which is capable of generating 1K resolution images. We validate our method by generating images at different resolutions, $2048 \times 2048$, $2048 \times 4096$, and $4096 \times 4096$. We used 50 EDM scheduler~\cite{karras2022elucidating} steps to generate images. We fixed our hyperparameters to noising step $\tau = 15$ and structural guidance step $\delta = 5$. We utilized Gaussian blur and sharpness factor $\alpha = 1.0$ for our sharpening operation. Hyperparameters are set equally in every experiment.

\subsection{Baselines}
We compare our method against two groups of baselines; training-free methods and super-resolution (SR) methods. For training-free methods, we selected (1) \textbf{ScaleCrafter}~\cite{he2023scalecrafter}, (2) \textbf{FouriScale}~\cite{huang2024fouriscale}, (3) \textbf{HiDiffusion}~\cite{zhang2023hidiffusion}, and (4) \textbf{DemoFusion}~\cite{du2024demofusion}. Each of these baselines are capable of generating higher-resolution images over the trained resolution with SDXL in a training-free manner. For SR methods, we compare our method to two popular SR models, namely (1) \textbf{SDXL+SD-Upscaler}~\cite{Rombach_2022_CVPR} and (2) \textbf{SDXL+BSRGAN}~\cite{zhang2021designing}, since it is intuitive to first generate an image and then apply super-resolution models to obtain higher-resolution images. In the main text, we mainly compare our method against training-free methods. Please refer the appendix for comparison to super-resolution methods.

\subsection{Evaluation}
We utilized the LAION-5B~\cite{schuhmann2022laion} dataset as a benchmark for the image generation experiments. Following previous works~\cite{du2024demofusion}, we randomly sampled 1K captions and generated images corresponding to each caption. We selected Frechet Inception Distance ($\text{FID}_r$)~\cite{heusel2017gans}, Kernel Inception Distance ($\text{KID}_r$)~\cite{binkowski2018demystifying}, and CLIP Score~\cite{radford2021learning} as our evaluation metrics. Note that $\text{FID}_r$ and $\text{KID}_r$ require resizing the images to a resolution of $299^2$, which is undesirable for assessing the high-frequency details of the image. To further provide the concrete evaluation, we also adopted patch FID ($\text{FID}_p$)~\cite{chai2022any} and patch KID ($\text{KID}_p$) as our evaluation metrics. In detail, we randomly cropped 1K patches from each generated image and measured the performance with randomly sampled 10K images from the LAION-5B dataset. For fair comparison between our proposed method and baselines, we ran the official code of each baseline and obtained the restuls.

\subsection{Results}

\begin{figure}[ht!]
    \centering
    \includegraphics[width=0.48\textwidth]{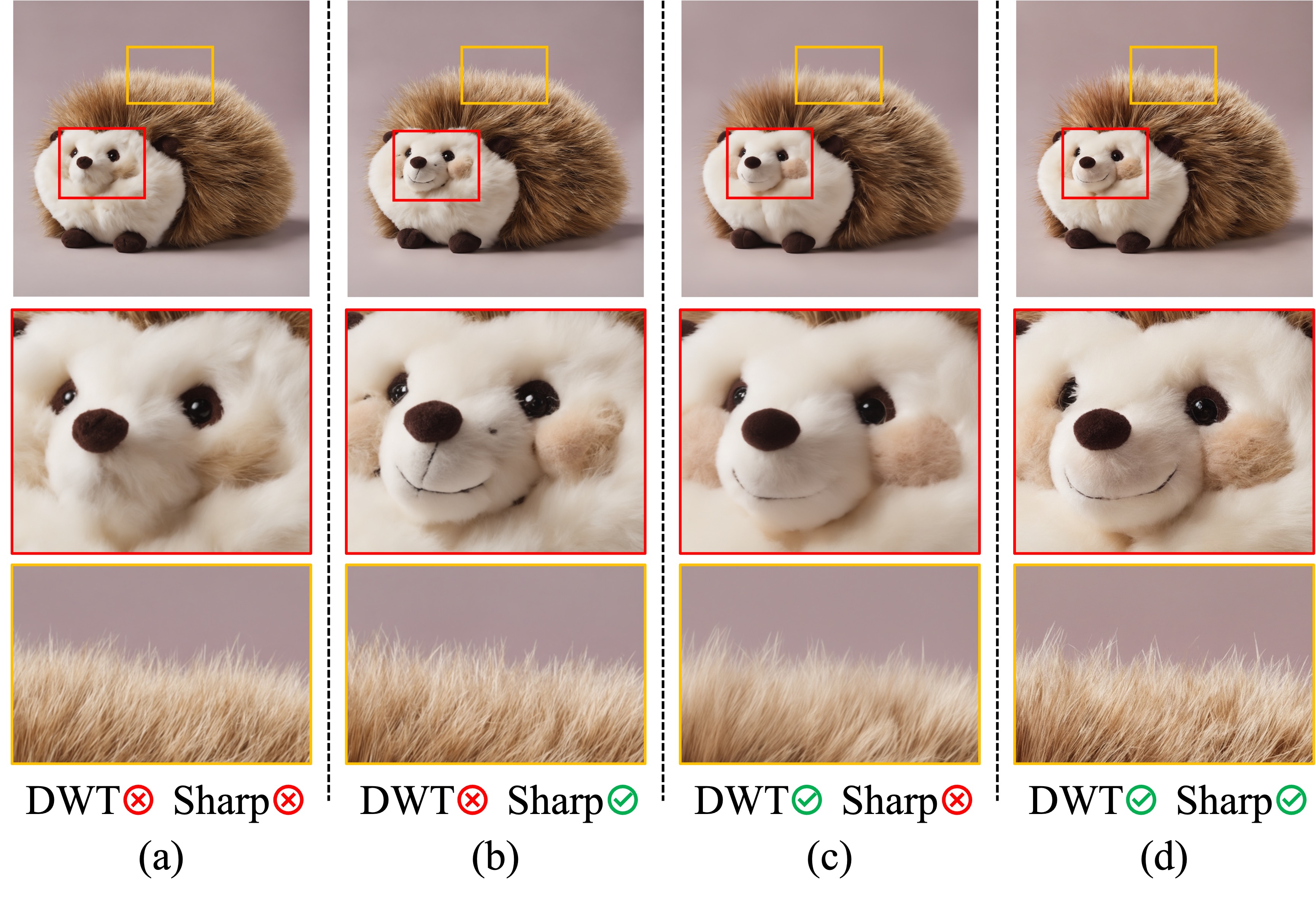} 
    \caption{\textbf{Ablating each component of \textit{DiffuseHigh}}. `DWT' denotes the DWT-based structural guidance and `Sharp' denotes the sharpening operation. Each sample has 4K resolution, generated from the same 1K image.} 
    \label{fig:fig_ablation1}
    \vspace{-10pt}
\end{figure}

\subsubsection{Qualitative Comparison}
We compare our method to baselines qualitatively in Fig.~\ref{fig:fig_comparison_with_baselines}. In terms of training-free methods, while ScaleCrafter, FouriScale, and HiDiffusion partially alleviate the object repetition problem, they often fail to capture correct global semantics, particularly at higher resolution. We argue that since these methods alter the pre-trained weights or the architecture of the UNet, they might have the risk of ruining the powerful generation ability of the diffusion model at higher resolution. DemoFusion preserves the overall structure of the image well. However, their approach frequently introduces small repeated objects in the local area of the result image, and also require considerable inference time, due to their MultiDiffusion~\cite{bar2023multidiffusion}-style generation scheme. Leveraging structural properties of the low-resolution images, \textit{DiffuseHigh} exhibits correct global structures, while also showcasing favorable textures and high-frequency details. 

\subsubsection{Quantitative Comparison}
We report the quantitative evaluation results in Tab.~\ref{tab:tab_main}. As observed, our method surpassed nearly every training-free baseline method in every resolution experiment, in terms of $\text{FID}_r$, $\text{KID}_r$, $\text{FID}_p$, and $\text{KID}_p$. These results demonstrate that our proposed \textit{DiffuseHigh} not only synthesizes visually approving results but also favorable textures and patterns corresponding to the higher-resolution images. One notable observation is that our metric scores does not differ a lot along the different resolutions compared to others, which proves the efficacy of our proposed method to transfer correct structures. Also, our method showcased superior performance on CLIP score, which highlights the ability of our pipeline to generate semantically correct images given text prompts. DemoFusion showed a better CLIP score compared to ours in the 2K and 4K experiments, but the difference is negligible. Moreover, our method achieved superb inference time thanks to our partial denoising process, which starts denoising process from the intermediate diffusion timestep.

\subsection{Ablation Studies}
\subsubsection{Structural Guidance and Sharpening} 
We validate the role of each component involved in our pipeline. As illustrated in Fig.~\ref{fig:fig_ablation1}, our structural guidance enables the generated image to preserve essential structures. By forcing the denoising process to maintain the low-frequency details of the sample, which is obtained from well-structured low-resolution images, samples with our DWT-based structural guidance present desirable structures and shapes. However, samples without structural guidance tend to have deformed shapes (mouth of the hedgehog in Fig.~\ref{fig:fig_ablation1} (a)) or artifacts (dots around the face in Fig.~\ref{fig:fig_ablation1} (b)). Also, we observed that the sharpening operation involved in our pipeline further enhances the quality of the image, particularly on blurred object boundaries or smoothed textures of the image (Fig.~\ref{fig:fig_ablation1} (c) and (d)). We also leave quantitative results in the appendix.

\begin{table}[ht!]
    \centering
    \resizebox{0.48\textwidth}{!}{
        \begin{tabular}{c|c c c c c c}
            \toprule
                $\delta$ & $\text{FID}_r (\downarrow)$ & $\text{KID}_r (\downarrow)$ & $\text{FID}_p (\downarrow)$ & $\text{KID}_p (\downarrow)$ & CLIP $(\uparrow)$ & Time\\
            \bottomrule
                3  & 56.35 & 0.007 & \textbf{38.67} & 0.010 & 31.31 & 223 sec\\
                5  & \textbf{56.09} & 0.007 & \underline{38.93} & 0.010 & \underline{31.32} & 258 sec \\
                7  & 56.18 & 0.007 & 38.95 & 0.010 & \textbf{31.33} & 293 sec \\
                10 & 56.49 & 0.007 & 39.31 & 0.011 & \underline{31.32} & 346 sec \\
                15 & 56.54 & 0.007 & 39.50 & 0.011 & 31.26 & 435 sec \\
            \hline
        \end{tabular}}
    \caption{\textbf{Evaluation with varying $\delta$}. We generated 10K images with randomly sampled captions from the LAION-5B dataset. We generated 4K images starting from the same 1K images generated by SDXL to ensure the fair comparison.}
    \label{tab:tab_ablation2}
    \vspace{-10pt}
\end{table}

\subsubsection{DWT-based Structural Guidance Steps}
We conduct the experiment with varying $\delta$ to assess the validity of our proposed structural guidance. As shown in Tab.~\ref{tab:tab_ablation2}, $\text{FID}_r$ decreases as $\delta$ approaches 5, and then increases as $\delta$ gets large. This observation suggests that our proposed structural guidance effectively facilitates the preservation of the desired structures, while an excessive guidance steps inhibit the generation of rich high-frequency details. Additionally, in terms of $\text{FID}_p$, $\delta = 3$ yielded the highest score and $\delta = 5$ the second highest score, but the difference is negligible. Nevertheless, we observed that small $\delta$ often fail to guide the correct global semantics. Therefore, we selected $\delta = 5$ as the optimal hyperparameter throughout this paper.


\section{Limitation and Discussion}
Since \textit{DiffuseHigh} leverages generated low-resolution images as structural guidance, the generation ability of the diffusion model at its original resolution heavily affects the overall performance of our method. That is, several structural defects or flaws in low-resolution images are also likely to be guided to the resulting higher-resolution image (please refer the examples in the appendix). However, we believe that leveraging tuning-free enhancement methods such as FreeU~\cite{si2023freeu}, which improve the generation ability of the diffusion model, would further improve the quality and fidelity of the resulting high-resolution image and leave it as a future work.

\section{Conclusion}
We present a training-free progressive high-resolution image synthesis pipeline using a pre-trained diffusion model. Our proposal involves leveraging generated low-resolution images as a guiding mechanism to effectively preserve the overall structure and intricate details of the contents. We also propose a novel principled way of incorporating structure information into the denoising process through frequency domain representation, which allows us to retain the essential information presented in low-resolution images. The extensive experiments with the pre-trained SDXL have shown that the proposed \textit{DiffuseHigh} generates higher-resolution images without commonly reported issues in the existing approaches, such as repetitive patterns and irregular structures.

\bibliography{aaai25}

\clearpage
\appendix
\setcounter{equation}{0}

\section{Preliminary}
We present preliminaries relevant to our method, including Latent Diffusion Models (LDM)~\cite{rombach2022high, podell2023sdxl} and Discrete Wavelet Transform (DWT).

\subsection{Latent Diffusion Model}
LDM is a diffusion model where the diffusion process is performed on a low-dimensional latent space. Given a data sample $x_0$ from the unknown data distribution $p_\text{data}(x)$, LDM encodes $x_0$ into a latent representation $z_0=\mathcal{E}(x_0)$, where $\mathcal{E}(\cdot)$ is an encoder that compresses the high-dimensional data into a compact latent space. 

Following the continuous-time DM framework~\cite{karras2022elucidating}, let $p(z; \sigma)$ be the latent distribution obtained by adding i.i.d Gaussian noise of variance $\sigma^2$ to the latent of the data. With sufficiently large $\sigma_{\text{max}}$, $p(z; \sigma_{\text{max}})$ is indistinguishable with pure Gaussian noise of variance $\sigma_{\text{max}}^2$, i.e., $p(z; \sigma_{\text{max}}) \approx \mathcal{N}(0, \sigma_{\text{max}}^2I)$. Initiating from $z_T \sim \mathcal{N}(0, \sigma_{\text{max}}^2I)$, DMs generate clean sample via solving the following stochastic differential equation (SDE):

\begin{equation}
    \begin{aligned}
    \label{eq:eq1_appendix}
        dz = & - \dot{\sigma}(t)\sigma(t) \nabla_z \log p(z; \sigma(t))dt \\
             & - \beta(t)\sigma(t)^2 \nabla_z \log p(z; \sigma(t)) dt \\
             & + \sqrt{2\beta(t)}\sigma(t) d\omega_t,
    \end{aligned}
\end{equation}
where $\omega_t$ is the standard Brownian motion and $\dot{\sigma}(t)$ the time-derivative of $\sigma(t)$. The solution of eq.~(\ref{eq:eq1_appendix}) can be found by numerical integration, which requires finite discrete sampling timesteps.

Consider the diffusion process with $T + 1$ timesteps. Defining the variances at each timestep as $0 = \sigma_0 < ... < \sigma_T = \sigma_\text{max}$, the denoising network $s_\phi(z_t; \sigma_t)$ parametrized by $\phi$ learns to estimate the score function $\nabla_{z} \log p(z; \sigma_t)$, which can be parametrized as follows:
\begin{equation}
    s_\phi(z_t; \sigma_t) = (D_\phi(z_t; \sigma_t) - z_t) / {\sigma_t}^2,
\end{equation}
where $z_t$ is a noisy latent at timestep $t$.

$D_\phi(\cdot)$ is a denoiser function that predicts the clean sample point given the noisy sample $z_t$, optimized via \textit{denoising score matching} objective:

\begin{equation}
    \mathbb{E}_{z_0, \epsilon} [\vert\vert D_\phi(z_t; \sigma_t) - z_0 \vert\vert_2^2],
\end{equation}
where
\begin{equation}
    z_t = z_0 + \epsilon, \quad \epsilon \sim \mathcal{N}(0, \sigma_t^2I).
\end{equation}

\subsection{Discrete Wavelet Transform}
\label{subsec:dwt}
Frequency-based methods, including the Discrete Fourier Transform (DFT), Discrete Cosine Transform (DCT), and Discrete Wavelet Transform (DWT) play a pivotal role in discrete signal processing. Such frequency-based approaches transform the given signal into the frequency domain, enabling the analysis and manipulation of the individual frequency bands.

Among them, utilizing wavelets, two-dimensional DWT decomposes images into different components that are localized both in time and frequency. Formally, let $\varphi$ be the two-dimensional scaling function, and $\psi^H, \psi^V, \psi^D$ the two-dimensional wavelets, each corresponding to the horizontal, vertical, and diagonal directions, respectively. Then, the single level 2D-DWT decomposition of the image $x$ can be written as:
\begin{equation}
    \texttt{DWT}(x) := \{W_\varphi(x)\} \cup \{W_{\psi^i}(x)\}_{i \in \{H,V,D\}},
\end{equation}

$W_\varphi(x)$ represents the approximation coefficient, and $W_{\psi^i}(x)$ the detail coefficients along the direction $i \in \{H, V, D\}$. Leveraging the low-pass filter and high-pass filter in both vertical and horizontal directions, the approximation coefficient $W_\varphi(x)$ represents the low-frequency details of the image, encompassing global structures, uniformly-colored regions, and smooth textures. On the other hand, the detail coefficients $W_{\psi^i}(x)$ encapsulates the high-frequency details, such as edges, boundaries, and rough textures.
\begin{figure*}[ht!]
    \centering
    \includegraphics[width=1.0\textwidth]{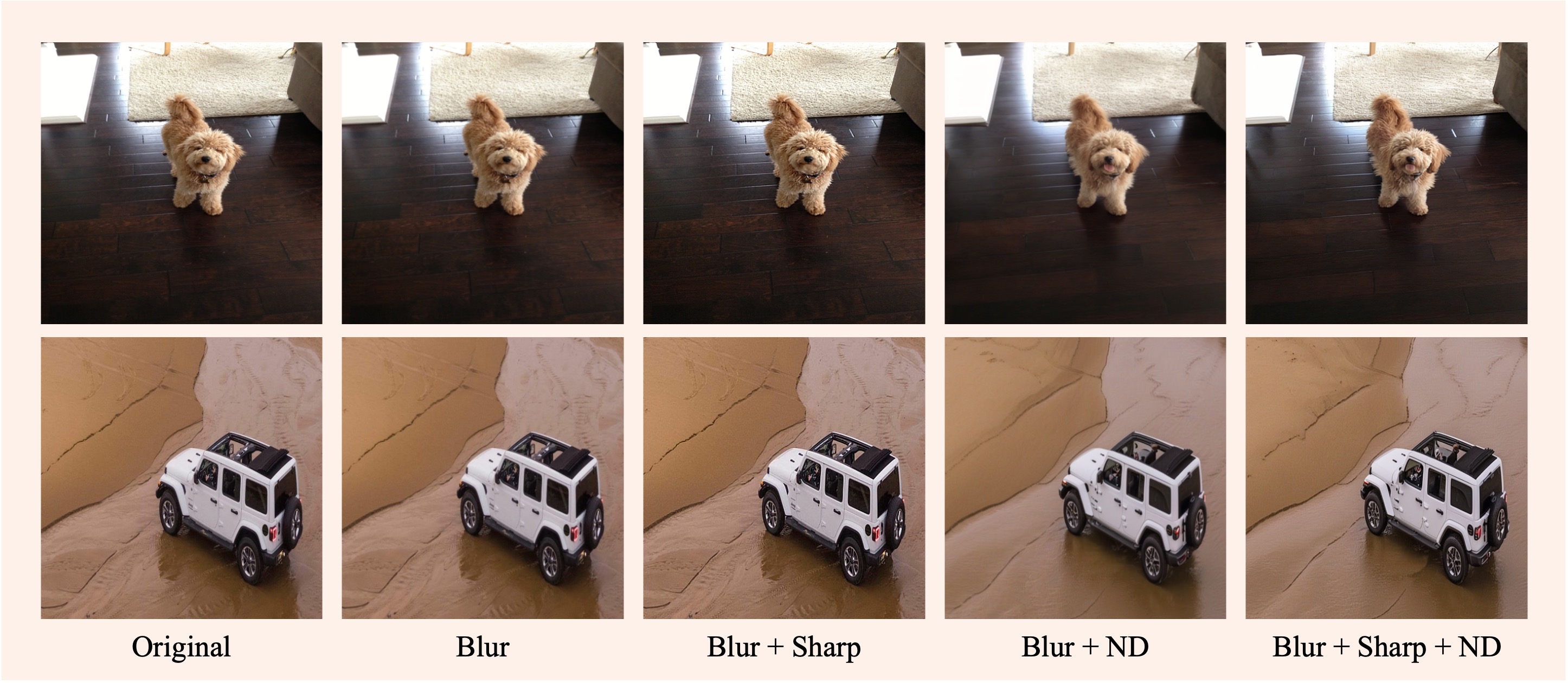} 
    \caption{\textbf{Toy experiment examples from the testset.} `ND' refers to Noising-Denoising process. Sharpening the blurry image, we can obtain more sharp and clean image after the noising-denoising process. Best viewed \textbf{ZOOMED-IN}.} 
    \label{fig:fig_toy_example}
\end{figure*}

\section{Toy Experiment on Sharpening Operation}
\begin{figure}[ht]
    \centering
    \includegraphics[width=0.48\textwidth]{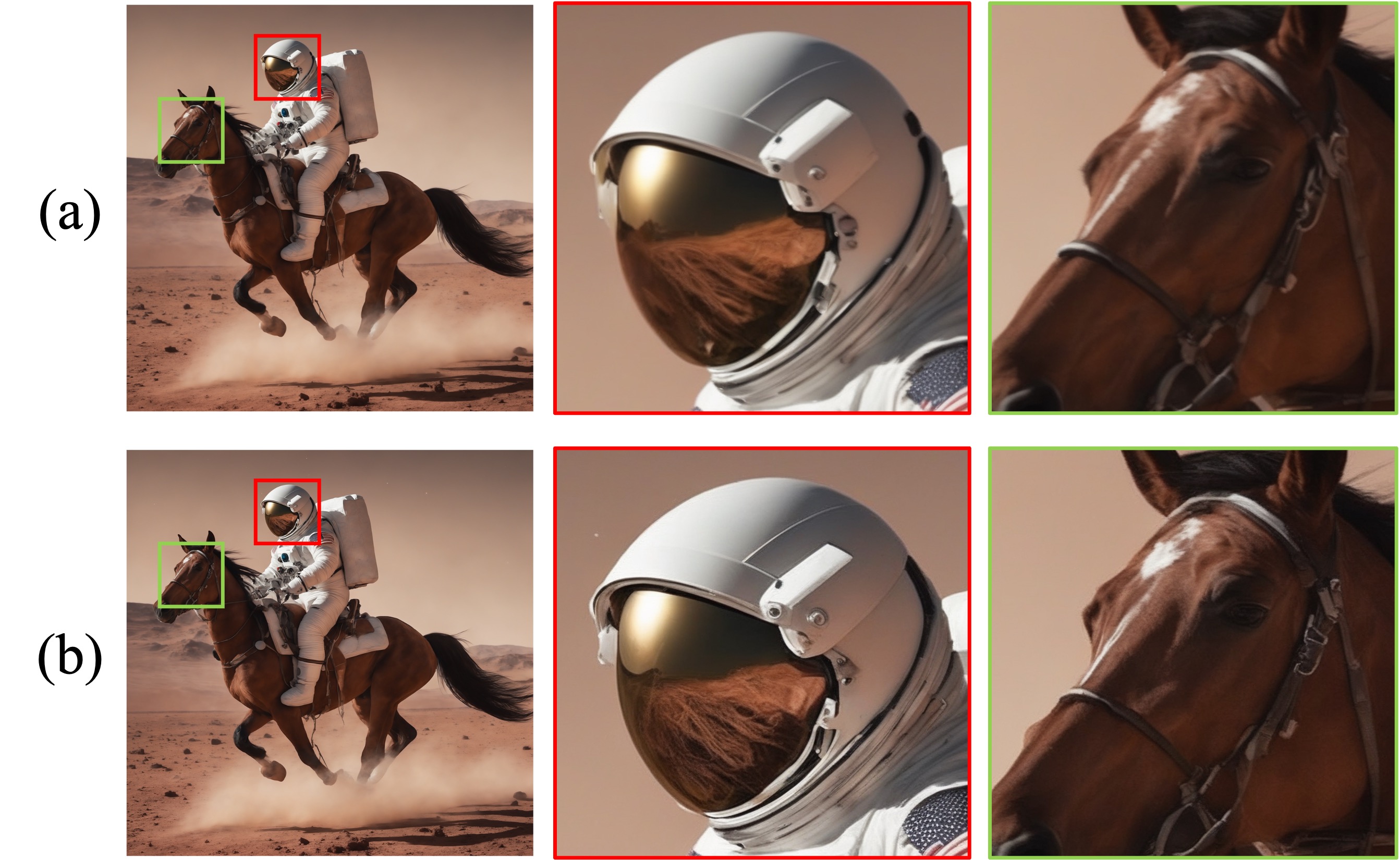} 
    \caption{\textbf{Effect of sharpening operation}. (a) \textit{DiffuseHigh} w/o sharpening, (b) \textit{DiffuseHigh} w/ sharpening. Incorporating sharpening operation, the generated sample shows more clear object boundaries and detailed textures. Each sample has $4096 \times 4096$ resolution.} 
    \label{fig:fig_comparison_sharp}
\end{figure}

As demonstrated in the main paper, we observed that involving sharpening operation in our pipeline successively alleviates the blurriness issue (see Fig.~\ref{fig:fig_comparison_sharp}), which arises from interpolating the low-resolution image. We constructed a toy experiment to study the denoising behavior of the SDXL, given noisy blurry image (noise added to the blurry image) and noisy sharp image (noise added to the sharpened image) as input.

 We randomly sampled 10K images above $512 \times 512$ resolution from the LAION-5B dataset and resized to $1024 \times 1024$ resolution. Then, we preprocessed the images to make two blurry image datasets, by (1) applying Gaussian blur to the image, and (2) downsampling and upsampling the images in order. Also, we applied the sharpening operation to the copy of each blurry dataset and obtained two sharpened image datasets. Finally, we added the Gaussian noise corresponding to the timestep $\tau$ to each dataset, and consecutively denoised the images with pre-trained SDXL. For evaluation, we measure the $\text{FID}_r$~\cite{heusel2017gans} and $\text{IS}_r$~\cite{salimans2016improved} score of each dataset. Additionally, we also report image Entropy~\cite{shannon1948mathematical} and mean variance of Laplacian (mVoL)~\cite{jain1995machine} to further evaluate the degree of sharpness of each denoised dataset. 
 
\begin{table}[ht!]
    \centering
    \resizebox{0.48\textwidth}{!}{
        \begin{tabular}{c|c c |c c}
            \toprule
                Dataset & $\text{FID}_r (\downarrow)$ & $\text{IS}_r (\uparrow)$ & Entropy $(\uparrow)$ & mVoL $(\uparrow)$ \\
            \bottomrule
                Gaussian Blur & 3.28 & 22.07 & 5.6198 & 98.15 \\
                Gaussian Blur + Sharp & \textbf{2.70} & \textbf{22.56} & \textbf{5.6455} & \textbf{337.85} \\ \hline
                DownUp  & 3.07 & 22.13 & 5.6177 & 56.67 \\
                DownUp + Sharp & \textbf{2.61} & \textbf{23.28} & \textbf{5.6382} & \textbf{210.66} \\
            \hline
        \end{tabular}}
    \caption{\textbf{Toy experiment results.} We added noise corresponding to $\tau$ to each blurry dataset and each sharpened dataset, and consecutively denoised with pre-trained SDXL.}
    \label{tab:tab_toy}
\end{table}

 We report the quantitative results of toy experiment in Tab.~\ref{tab:tab_toy}. As demonstrated, incorporating a sharpening operation achieved better $\text{FID}_r$ and $\text{IS}_r$ scores in all cases, indicating that sharpening enhances the recovery of desirable textures and details in blurry samples during the denoising process. Additionally, the sharpened dataset achieved higher Entropy and mVoL scores, reflecting improved sharpness and intensity changes in the result images. These observation support our claim that the sharpening operation helps the data samples to be located more closely to the sharp data distribution, facilitating a better alignmnet with the sharp data distribution mode after the noising-denoising process. We also provide selected testset examples in Fig.~\ref{fig:fig_toy_example}.

\begin{table}[ht!]
    \centering
    \resizebox{0.48\textwidth}{!}{
        \begin{tabular}{c c|c c|c c|c}
            \toprule
                DWT & Sharp & $\text{FID}_r (\downarrow)$ & $\text{FID}_p (\downarrow)$ & $\text{Entropy} (\uparrow)$& $\text{mVoL} (\uparrow)$ & CLIP $(\uparrow)$\\
            \bottomrule
                \ding{55} & \ding{55} & 56.49 & \underline{38.93} & 7.0386 & 10.42 & 31.31 \\
                \ding{55} & \ding{51} & 56.39 & \textbf{38.89} & \textbf{7.0633} & \underline{21.74} & \textbf{31.32} \\
                \ding{51} & \ding{55} & \underline{56.24} & 38.95 & 7.0449 & 11.22 & 31.31 \\
                \ding{51} & \ding{51} & \textbf{56.09} & \underline{38.93}& \underline{7.0624}  & \textbf{26.88} & \textbf{31.32} \\
            \hline
        \end{tabular}}
    \caption{\textbf{Quantitative results of ablating \textit{DiffuseHigh} components.} The experiment is performed on $4096 \times 4096$ image generation settings.}
    \label{tab:tab_ablation1}
\end{table}

\begin{figure*}[ht!]
    \centering
    \includegraphics[width=1.0\textwidth]{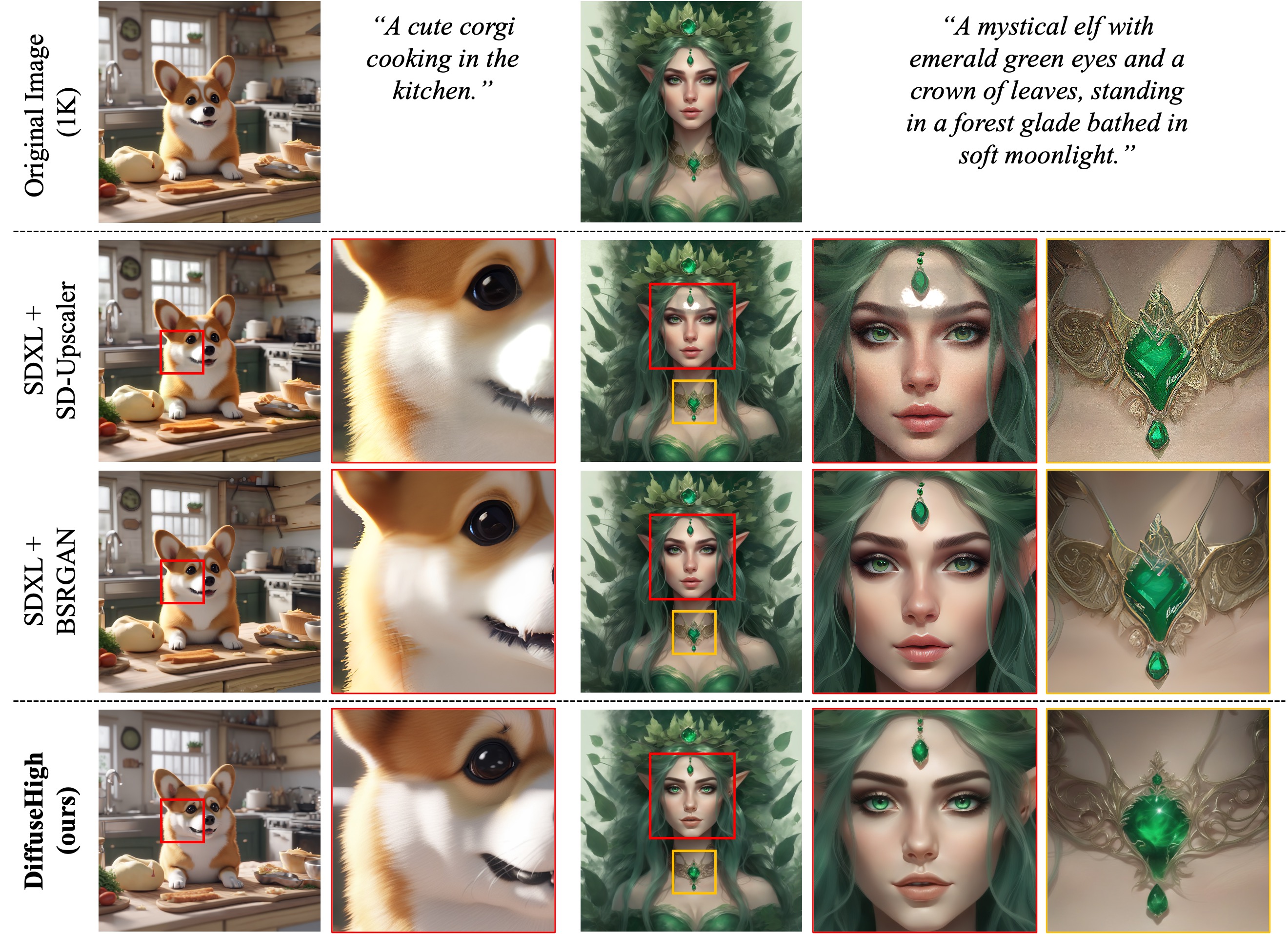} 
    \caption{\textbf{Qualitative comparison to SR models.} Each images have $4096 \times 4096$ resolution.} 
    \label{fig:fig_comparison_SR}
\end{figure*}

\begin{table*}[ht!]
    \centering
    \resizebox{\textwidth}{!}{
        \begin{tabular}{c|c c c c c|c c c c c|c c c c c}
            \toprule
                & \multicolumn{5}{c|}{$2048 \times 2048$} & \multicolumn{5}{c|}{$2048 \times 4096$} & \multicolumn{5}{c}{$4096 \times 4096$} \\ \hline
                
               Methods & $\text{FID}_r (\downarrow)$ & $\text{KID}_r (\downarrow)$ & $\text{FID}_p (\downarrow)$ & $\text{KID}_p (\downarrow)$ & CLIP $(\uparrow)$ & $\text{FID}_r (\downarrow)$ & $\text{KID}_r (\downarrow)$ & $\text{FID}_p (\downarrow)$ & $\text{KID}_p (\downarrow)$ & CLIP $(\uparrow)$ & $\text{FID}_r$ & $\text{KID}_r (\downarrow)$ & $\text{FID}_p (\downarrow)$ & $\text{KID}_p (\downarrow)$ & CLIP $(\uparrow)$ \\

            \midrule
               SDXL+SD-Upscaler & 59.58 & \textbf{0.008} & 38.47 & 0.012 & \underline{32.21} & 58.27 & \underline{0.009} & \underline{38.17} & \underline{0.013} & \underline{31.19} & 56.67 & \textbf{0.006} & \underline{44.73} & \underline{0.014} & \underline{32.41} \\
               SDXL+BSRGAN & \textbf{55.25} & 0.006 & \textbf{33.40} & \textbf{0.008} & \textbf{32.51} & \underline{57.86} & \underline{0.009} & 40.63 & 0.019 & \textbf{31.20} & \textbf{55.52} & \textbf{0.006} & 46.48 & 0.018 & \textbf{32.47} \\
            \bottomrule
               DiffuseHigh & \underline{56.90} & \textbf{0.008} & \underline{34.14} & \underline{0.009} & 31.19 & \textbf{53.83} & \textbf{0.007} & \textbf{33.73} & \textbf{0.009} & 30.50 & \underline{56.09} & 0.007 & \textbf{38.93} & \textbf{0.010} & 31.32 \\
            \hline
        \end{tabular}}
    \caption{\textbf{Quantitative results of camparison to SR methods}. We ran the official code of each baselines and obtained results.}
    \label{tab:tab_SR}
\end{table*}

\begin{figure*}[ht!]
    \centering
    \includegraphics[width=1.0\textwidth]{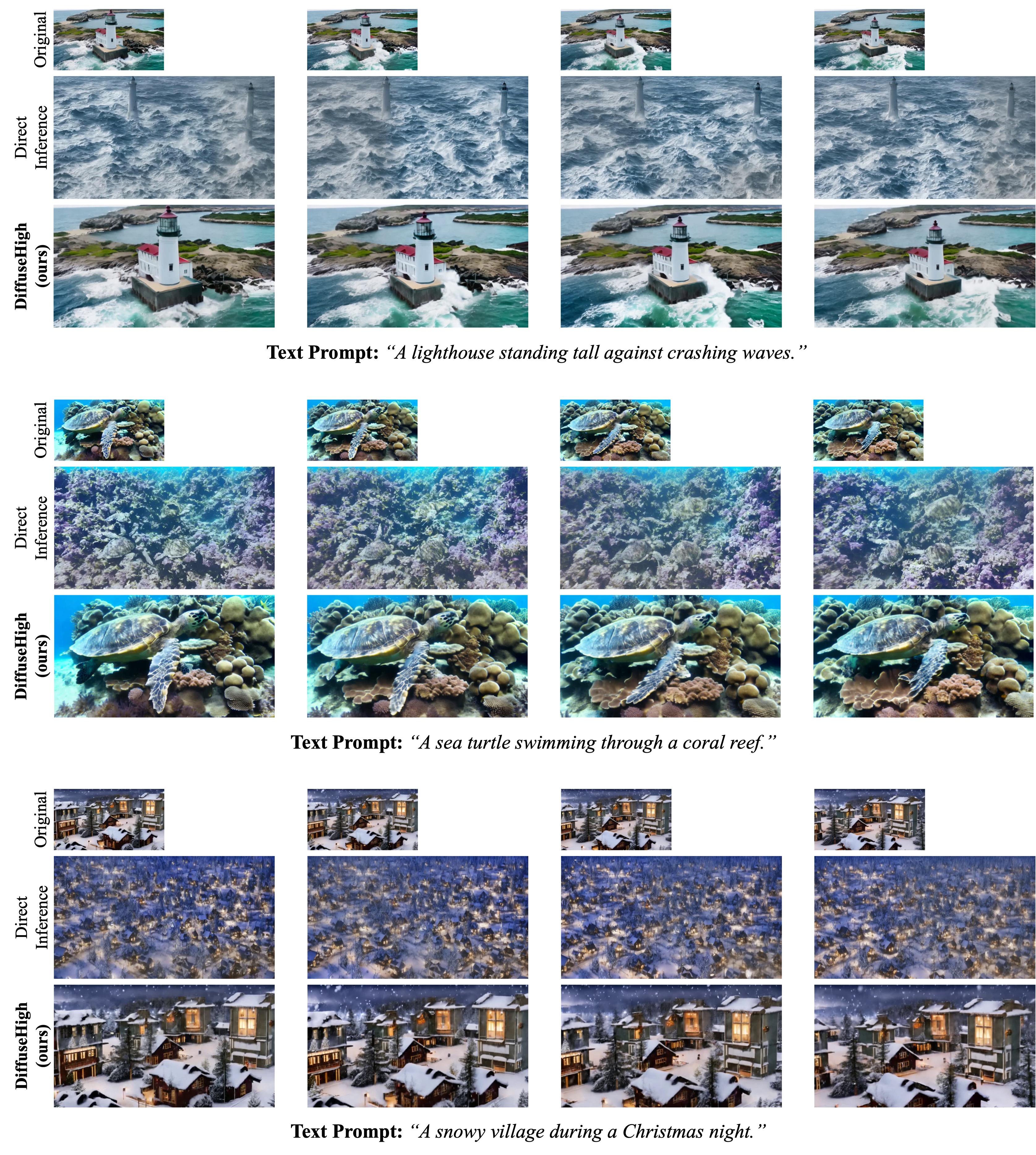} 
    \caption{\textbf{Applying \textit{DiffuseHigh} on ModelScope.} We generate videos with $4\times$ higher resolution, $640 \times 1152$ resolution, than the original resolution, i.e., $320 \times 576$ resolution videos.} 
    \label{fig:fig_modelscope}
\end{figure*}

\section{Quantitative Results of Ablation Study}

We present the quantitative results of ablating each component of \textit{DiffuseHigh} in Tab.~\ref{tab:tab_ablation1}. To assess performance, we generated 4K resolution images from the same 1K resolution inputs for $\text{FID}_r$, and cropped the same region from each image for $\text{FID}_p$. We observed that $\text{KID}_r$ and $\text{KID}_p$ did not show notable difference between each method, therefore alternatively report Entropy and mVoL score, in order to compare the sharpness of the result images. As observed, combining both DWT-based structural guidance and sharpening operations yields the best $\text{FID}_r$ score. While the sharpening operation alone produced the best $\text{FID}_p$ score, the differences between the methods are negligible. We argue that since $\text{FID}_p$ evaluates only local patches of the image, it lacks the capacity to determine whether small artifacts or flaws (as shown in Fig. 5 (b) of our main text) represent correct texture or not. In terms of sharpness metrics, e.g, Entropy and mVoL, method with sharpening operation always yield higher score. For a balanced performance across all evaluation metrics, we found it advantageous to utilize every component of our \textit{DiffuseHigh} pipeline.

\section{Comparison to Super-Resolution Models}
We compare our proposed \textit{DiffuseHigh} against SR models both qualitatively and quantitatively. Note that SR models demands a large number of high-resolution images and substantial computational resources for the training, whereas our method operates in a completely training-free manner. Qualitative results are shown in Fig.~\ref{fig:fig_comparison_SR}. As shown, results with \textit{DiffuseHigh} exhibit more detailed and appropriate details and textures. While SR methods effectively preserve the correct structures from low-resolution image inputs, they tend to produce simply smoothed images, without sufficient high-frequency details. This observation alerts us that directly utilizing pre-trained SR models may not guarantee the appropriate injection of high-frequency details in higher resolution image generation settings.

We report quantitative results comparing our pipeline to SR models in Tab.~\ref{tab:tab_SR}. Notably, even though our proposed \textit{DiffuseHigh} does not require any training or fine-tuning, our method demonstrates comparable performance to SR methods. Generally, SR models acheive slightly better scores on $\text{FID}_r$ and $\text{KID}_r$, since SR models are designed to align precisely with the low-resolution image, and these metrics require resizing the result image into the low-resolution input size. However, our method outperforms SR models on $\text{FID}_p$ and $\text{KID}_p$, particularly at higher resolutions ($2048 \times 4096$ and $4096 \times 4096$), indicating that our approach is better suited for introducing appropriate high-frequency details into the image.

\begin{figure*}[ht!]
    \centering
    \includegraphics[width=0.9\textwidth]{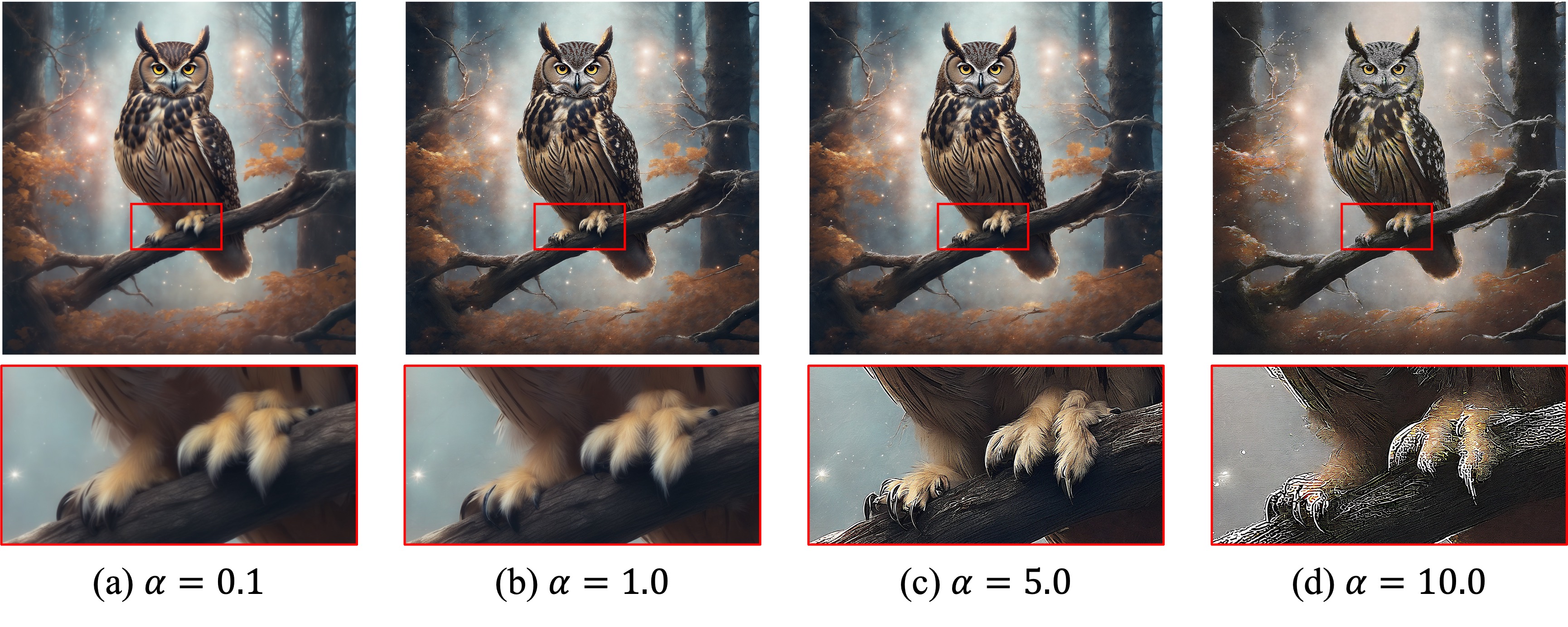} 
    \caption{\textbf{Varying sharpness factor $\alpha$.} Image generated with small $\alpha$ still produces blurry texture (See (a)), while images with large $\alpha$ shows severe artifacts (See (c) and (d)). } 
    \label{fig:fig_sharpness_factor}
\end{figure*}

\begin{figure}[ht!]
    \centering
    \includegraphics[width=0.5\textwidth]{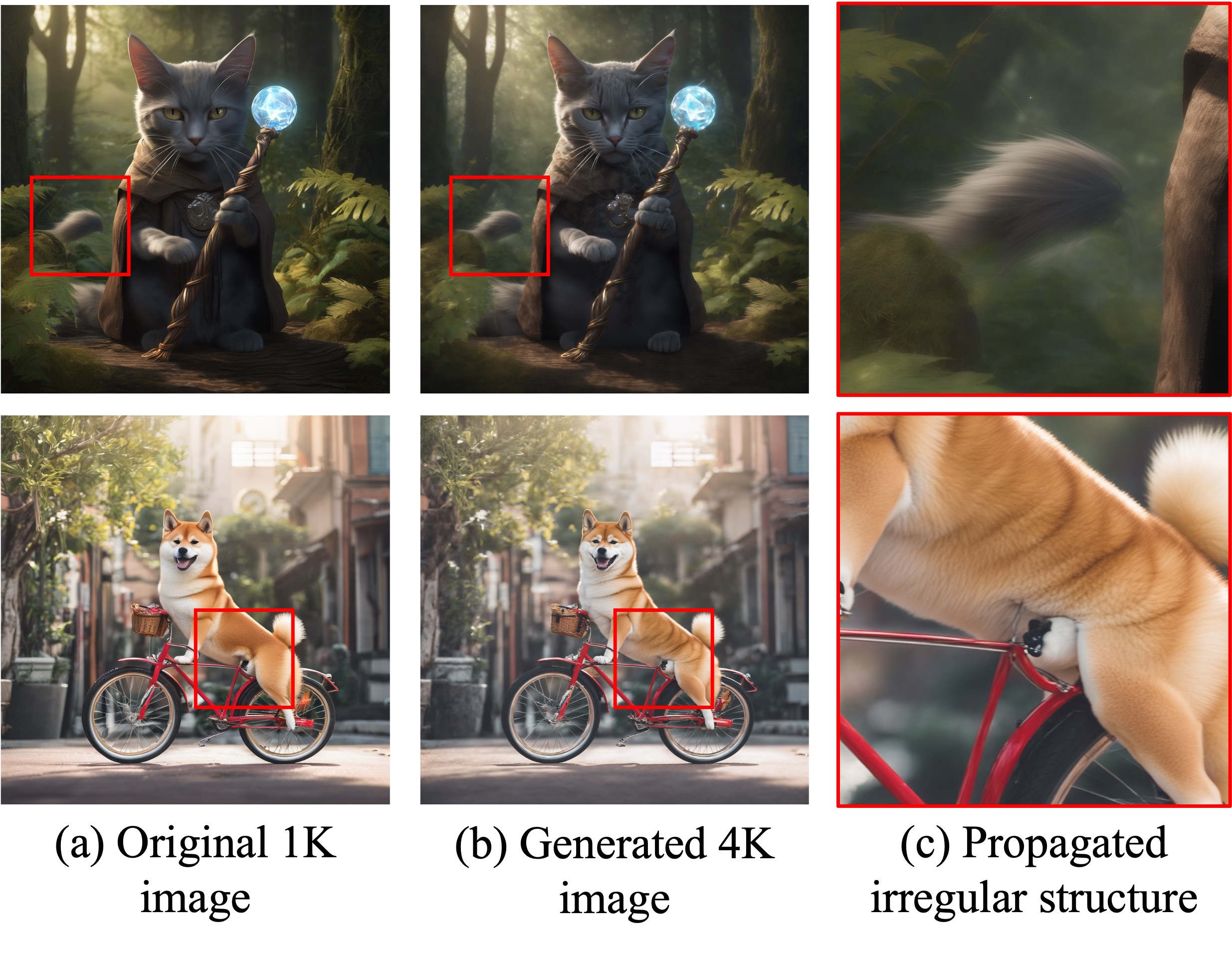} 
    \caption{\textbf{Failure cases of \textit{DiffuseHigh}.} Structural flaws originating from the low-resolution images are also present in the generated high-resolution images.} 
    \label{fig:fig_failure_case}
\end{figure}

\section{Applying \textit{DiffuseHigh} to the Text-to-Video Diffusion Models}
Our proposed \textit{DiffuseHigh} does not modify the pre-trained weight of the diffusion model, thus easily applied to the other diffusion models. We observed that directly infererring the text-to-video diffusion model at a higher resolution than its training resolution also leads to issues such as object repetition and irregular structures. Here, we show that \textit{DiffuseHigh} can be successively adapted to the text-to-video diffusion model, showcasing the versatility of our method.

We utilized ModelScope~\cite{wang2023modelscope}, a text-to-video model capable of generating videos at a resolution of $320 \times 576$. We generated videos at $4\times$ higher resolution, i.e., $640 \times 1152$, using both vanilla ModelScope and ModelScope with \textit{DiffuseHigh}. As shown in Fig.~\ref{fig:fig_modelscope}, direct inference of ModelScope at a resolution higher than its training resolution suffers from repeated objects and chaotic patterns. By simply applying our pipeline, ModelScope successively generates higher-resolution videos with correct structures and improved details.

\section{Varying Sharpness Factor $\alpha$}
Over-sharpening an image typically suffers from increased noise, color shifts, and a loss of details. We found that these drawbacks also occur in our pipeline when we adopt too large $\alpha$, as illustrated in Fig.~\ref{fig:fig_sharpness_factor}. Empirically, we found that setting $\alpha = 1.0 \sim 2.0$ works well in general cases.

\section{Failure Cases}
We report the failure case of our \textit{DiffuseHigh} in Fig.~\ref{fig:fig_failure_case}. As shown, our DWT-based structural guidance inevitably guides the incorrectly synthesized obejcts (Fig.~\ref{fig:fig_failure_case} first row) or structural flaws (Fig.~\ref{fig:fig_failure_case} second row) into the generated high-resolution image, originated from the low-resolution image.

\section{More Qualitative Results}
\begin{figure*}[ht!]
    \centering
    \includegraphics[width=0.93\textwidth]{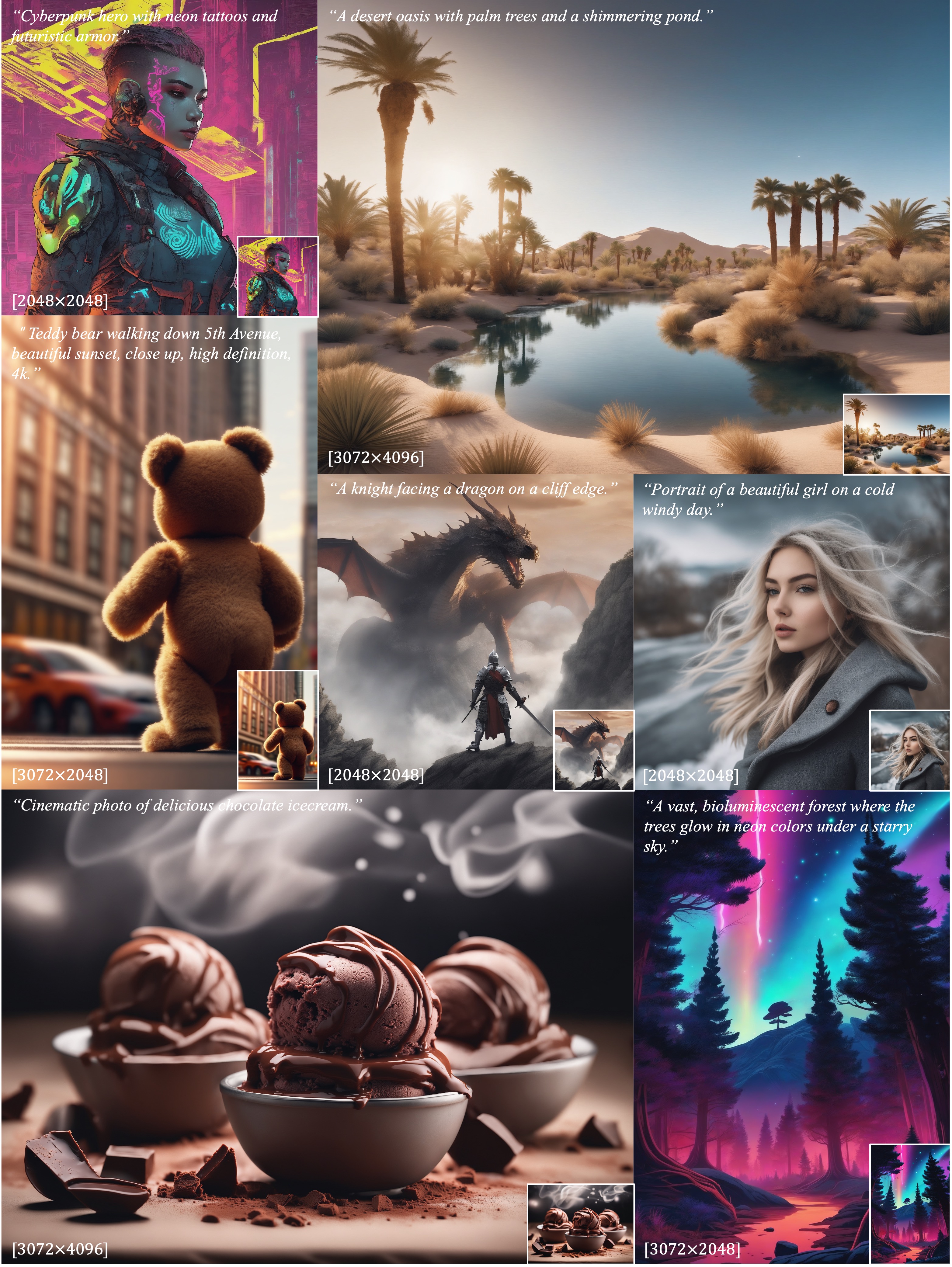} 
    \caption{\textbf{More qualitative results of \textit{DiffuseHigh}.}} 
    \label{fig:fig_cherry_1}
\end{figure*}

\begin{figure*}[ht!]
    \centering
    \includegraphics[width=0.9\textwidth]{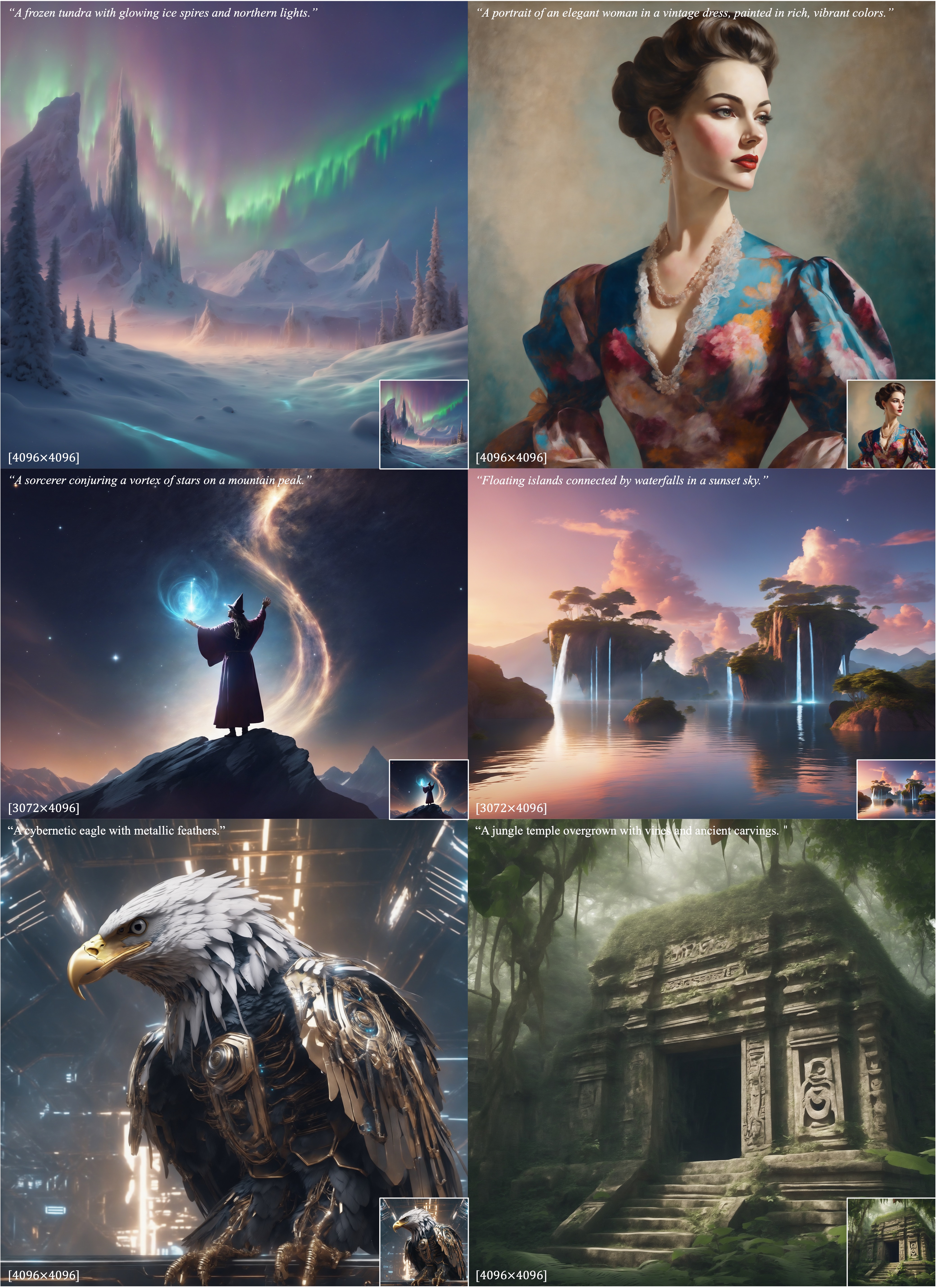} 
    \caption{\textbf{More qualitative results of \textit{DiffuseHigh}.}} 
    \label{fig:fig_cherry_2}
\end{figure*}

We provide more qualitative samples generated with \textit{DiffuseHigh} in Fig.~\ref{fig:fig_cherry_1} and Fig.~\ref{fig:fig_cherry_2}. We also provide original resolution sample in bottom right corner of each samples.

\end{document}